\documentclass[a4paper,11pt]{article}
\usepackage[margin=1in]{geometry}

\usepackage[colorlinks,linkcolor=blue,citecolor=blue]{hyperref}
\usepackage{amsmath,amssymb,amsthm}
\usepackage[round]{natbib}
\usepackage{mathabx}
\usepackage{graphicx} 
\usepackage{subfigure} 
\usepackage{algorithm,algorithmic}
\usepackage{tabularx}
\usepackage{enumitem}

\usepackage{amsmath,amsfonts,amssymb}
\usepackage{amsthm} 

\usepackage[capitalise]{cleveref}
\crefname{figure}{Figure}{Figures}

\usepackage[table]{xcolor}
\usepackage{tikz}
\usetikzlibrary{arrows,shapes,arrows.meta}

\newcommand{\ignore}[1]{}

\theoremstyle{plain}
\newtheorem{theorem}{Theorem}
\newtheorem{lemma}[theorem]{Lemma}

\newtheorem*{theorem*}{Theorem}
\newtheorem*{lemma*}{Lemma}
\newtheorem*{corollary*}{Corollary}
\newtheorem*{proposition*}{Proposition}
\newtheorem*{claim*}{Claim}
\newtheorem*{fact*}{Fact}

\theoremstyle{definition}

\newtheorem*{definition*}{Definition}

\newtheorem*{remark*}{Remark}

\newtheorem*{example*}{Example}

\theoremstyle{plain}
\newtheorem*{theoremaux}{\theoremauxref}
\gdef\theoremauxref{1}

\DeclareMathAlphabet{\mathbfsf}{\encodingdefault}{\sfdefault}{bx}{n}

\let\Pr\relax
\DeclareMathOperator{\Pr}{\mathbb{P}}

\newcommand{\lr}[1]{\!\left(#1\right)\!}

\newcommand{\lrbra}[1]{\!\left[#1\right]\!}

\newcommand{\set}[1]{\{#1\}}

\newcommand{\wt}[1]{\smash{\widetilde{#1}}}

\renewcommand{\O}{O}

\newcommand{\tO}{\wt{\O}}

\newcommand{\tOmega}{\wt{\Omega}}
\newcommand{\E}{\mathbb{E}}
\newcommand{\EE}[1]{\E\lrbra{#1}}

\newcommand{\N}{\mathcal{N}}

\newcommand{\reals}{\mathbb{R}}

\newcommand{\half}{\frac{1}{2}}

\let\nablaold\nabla
\renewcommand{\nabla}{\nablaold\mkern-2.5mu}

\newcommand{\KL}[2]{D_{\mathrm{KL}}\lr{#1 \,\middle\|\, #2}}
\newcommand{\Regret}{\mathrm{R}_T}
\newcommand{\TV}[2]{D_{\mathrm{TV}}\lr{#1 \,,\, #2}}
\newcommand{\clip}{\mathrm{clip}}

\newcommand{\stpath}{\ensuremath{s\textnormal{-}t}\ }

\title{Tight Bounds for Bandit Combinatorial Optimization}
\author{
Alon Cohen \qquad\quad Tamir Hazan\\
\normalsize{Technion---Israel Institute of Technology}\normalsize\\
\texttt{\normalsize \{alon.cohen,tamir.hazan\}@technion.ac.il} \\
\and
Tomer Koren\\
\normalsize{Google Brain}\\
\texttt{\normalsize tkoren@google.com}
}

\begin{document} 
\maketitle

\begin{abstract}
We revisit the study of optimal regret rates in bandit combinatorial optimization---a fundamental framework for sequential decision making under uncertainty that abstracts numerous combinatorial prediction problems.
We prove that the attainable regret in this setting grows as $\wt{\Theta}(k^{3/2}\sqrt{dT})$ where $d$ is the dimension of the problem and $k$ is a bound over the maximal instantaneous loss, disproving a conjecture of \cite*{audibert2014regret} who argued that the optimal rate should be of the form $\wt{\Theta}(k\sqrt{dT})$.
Our bounds apply to several important instances of the framework, and in particular, imply a tight bound for the well-studied bandit shortest path problem.
By that, we also resolve an open problem posed by \cite*{cesa2012combinatorial}.



\end{abstract}

\section{Introduction}

We consider the problem of online combinatorial optimization with bandit feedback, also known as \emph{bandit combinatorial optimization}, or more succinctly as \emph{combinatorial bandits}.
The problem can be described as the following game between a learner and an environment, that proceeds for $T$ rounds. 
On each round $t =1,2,\ldots,T$, the learner has to pick, possibly at random, an action $x_t$ from a subset $S \subseteq \set{0,1}^d$ of the hypercube in $d$-dimensions, with the property that each element $x \in S$ has at exactly $k$ non-zero entries, that is $\sum_{i=1}^d x_i = k$. 
Simultaneously, the environment privately chooses a loss vector $\ell_t \in [0,1]^d$.
The learner then incurs the loss $\ell_t \cdot x_t \in [0,k]$ and may observe only this loss (but not the vector $\ell_t$) as feedback.
%
%
%
The goal of the learner throughout the $T$ rounds of the game is to minimize her regret, defined as
\begin{equation*}
\sum_{t=1}^T \ell_t \cdot x_t ~-~ \min_{x \in S} \sum_{t=1}^T \ell_t \cdot x 
~.
\end{equation*}

Bandit combinatorial optimization is a fundamental primitive of sequential decision making under uncertainty, and abstracts several major problems in this context (see, e.g., \citealp{bubeck2012regret}).
Perhaps the most important and well-studied problem captured by this framework is online network routing, also known as the online shortest path problem \citep{takimoto2003path, kalai2005efficient}.
In this setting, 
a source station $s$ repeatedly sends communication packets to a target station $t$ through a network represented by a connected directed acyclic graph. 
On each decision round, the environment associates each edge in the network with a loss, and the learner suffers the loss accumulated over the edges in her chosen path. 
Each packet can be routed differently and the station has to pick routes so as to minimize the overall amount of time it takes the packets to arrive. 
In the bandit version of the problem, the only feedback that the source station observes is the roundtrip time of each packet---namely the time it takes the packet to travel to its destination and return to the source. 
%

The network routing problem can be cast in the online combinatorial optimization framework as follows: the set of all \stpath paths can be represented as a set $S \subseteq \set{0,1}^d$ where $d$ is the number of edges in the graph, and the non-zero entries in each $x \in S$ indicate the edges that are contained in the path $x$;
then, if $\ell_t \in [0,1]^d$ is the loss vector that associates costs to edges in the network on decision round $t$, then the cost of path $x$ is given by $\ell_t \cdot x$.
The assumption that $\sum_{i=1}^d x_i = k$ for all $x \in S$ means that the length of an \stpath path in the network is exactly $k$ (which is also an upper bound on the maximal cost of any \stpath path).

The study of bandit combinatorial optimization dates back to the work of
\cite{awerbuch2004adaptive}, who considered the online shortest path problem in the bandit setting, henceforth called the \emph{bandit shortest path} problem, in which the learner observes only the loss that she has suffered, and showed an $O(k d^{5/3} T^{2/3})$ bound on the expected regret.
%
%
\cite{dani2008price} and \cite{abernethy2008efficient} 
considered the problem in the wider context of bandit linear optimization and established a regret bound with the optimal $\sqrt{T}$ dependence.
Subsequently, \cite{cesa2012combinatorial} focused on bandit combinatorial optimization, and showed that a similar bound can be achieved for a large number of problems under this framework, often with computationally efficient algorithms. 
For the bandit shortest path problem, \cite{cesa2012combinatorial} conjectured that the general upper bound is in fact suboptimal and that the correct tight bound is of the form $\tO(k\sqrt{dT})$, and could be obtained by a clever adaptation of their algorithm.

More recently, \cite{audibert2014regret} showed that the aforementioned $\tO(k^{3/2}\sqrt{dT})$ upper bound holds for any combinatorial bandit problem using a general online optimization algorithm.
Additionally, the authors gave a new lower bound of~$\Omega(k \sqrt{dT})$ on the expected regret in combinatorial bandits, which leaves a gap of $\sqrt{k}$ between that and their upper bound (ignoring logarithmic factors). 
They conjectured as well that the lower bound is, in fact, the correct rate and articulated that the upper bound could be improved by non-trivial modifications of the existing algorithmic techniques.

In this paper, we revisit the study of optimal regret rates in bandit combinatorial optimization. 
Our main contribution is in disproving the conjectures of \cite{cesa2012combinatorial} and \cite{audibert2014regret}
and showing that the expected regret of combinatorial bandits in general, and of the bandit shortest path problem in particular, is in fact $\wt{\Theta}(k^{3/2} \sqrt{d T})$.
Namely, we show a new lower bound of $\tOmega(k^{3/2} \sqrt{d T})$ for combinatorial bandits that matches the best known upper bound up to logarithmic factors, and also holds (via simple adaptations) in the context of bandit shortest path.
%
Furthermore, we show how this lower bound can be adapted to the setting of online ranking \citep{helmbold2009learning}.

Surprisingly, the construction used in our lower bound is very simple and is based on straightforward adaptations of the one used by \cite{audibert2014regret}.  
Furthermore, our analysis is also significantly simpler and shorter than theirs.
In a nutshell, the improvement in the bound is obtained via the following observation: when picking its randomized losses for fooling the learner, the environment can choose noisy vectors whose entries are strongly correlated with each other rather than being independent, as is the case in typical lower bound constructions (and, in particular, as suggested by \citealp{audibert2014regret}).%
\footnote{Note that the correlation discussed here is between different entries of the same loss vector, rather than between different loss vectors at different rounds. In particular, the loss vectors in our lower bound constructions are still chosen i.i.d.~so our bounds also apply to the stochastic i.i.d.~case.}
Since the learner never observes individual entries of the loss vectors and can only see a sum of $k$ of them (for a particular choice of the action set $S$), she cannot exploit this correlation in any way. 
On the other hand, with correlated noise terms the observed loss value can have a variance that grows \emph{quadratically} with $k$, rather linearly as is the case with i.i.d.~noise, which directly deteriorates the learner's regret by an additional factor of $\sqrt{k}$.

\subsection{Related work}

Combinatorial bandit optimization is closely related to a somewhat more general online learning scenario known as \emph{bandit linear optimization}, which was first considered by \cite{dani2008price} and \cite{abernethy2008efficient}.
In this setting, the decision set $S$ is not restricted to subsets of the hypercube $\set{0,1}^d$ and may be an arbitrary compact convex set in $\reals^d$; instead, the only requirement is that the loss the learner incurs by picking any action in $S$ is bounded (say, by $1$ in absolute value) for all possible loss vectors of the environment.
State-of-the-art bounds for this problem were obtained by \cite{bubeck2012towards} and \cite{hazan2016volumetric}, the latter using computationally-efficient~algorithms.

The general linear optimization setting allows for more general geometries of the sets in which the decisions and the loss vectors reside (e.g., they are typically assumed to be subsets of the Euclidean unit ball), and consequently the bounds obtained in that setting are often not immediately comparable to those in the combinatorial one.
In particular, the lower bounds proved by \cite{dani2008price} and more recently by \cite{shamir2015complexity} hold in the general linear optimization setting (with Euclidean geometry) and do not apply to any natural problem in the combinatorial setting.



A significant amount of work has been devoted to combinatorial optimization in the closely related semi-bandit feedback model (e.g., \citealp{gyorgy2007line,kale2010non,audibert2014regret,neu2015first,neu2016importance}), in which after playing an action $x_t$ the learner may observe the individual entries of the loss vector $\ell_t$ that correspond to active entries of $x_t$, namely those entries~$i$~for which $x_t(i)=1$.
For example, in the context of the online shortest path problem, instead of observing just the overall cost of the chosen path (as is the case in the bandit setting), the player may observe the individual cost of each edge in that path.
In the semi-bandit case, however, the regret of bandit combinatorial optimization is by now well understood, and is known to be of the form $\Theta(\sqrt{kdT})$; see \cite{audibert2014regret} and the references therein.

For further and more detailed account on related partial information models and their regret analysis, we refer to the recent survey by \cite{bubeck2012regret}.

\section{Main results}
\label{sec:mainresults}
We now state the main results of this paper.
As our results are lower bounds on the learner's regret, we will 
henceforth focus on oblivious environments, that are required to choose the entire sequence $\ell_1,\ldots,\ell_T$ before the game begins and thus do not react adaptively to the player's randomized decisions. 
(A lower bound for such environments also implies a lower bound for more general adaptive environments.)
In this setup, we will give bounds on the expected regret, defined as
\begin{equation}
\label{eq:regret}
\Regret 
= 
\EE{ \sum_{t=1}^T \ell_t \cdot x_t } ~-~ \min_{x \in S} \sum_{t=1}^T \ell_t \cdot x 
~,
\end{equation}
where the expectations are taken over the random choices of the learner.

Our first result deals with the general combinatorial bandits setting and shows that if the environment is free to choose any action set $S$, the regret of the learner can be very large. 
Our lower bound is attained in the multitask bandit problem, in which a learner is simultaneously trying to solve $k$ instances of the $n$-armed bandit problem \citep{auer2002nonstochastic} with $n=d/k$ (we assume for simplicity that the latter is an integer). At every round of the game, the learner plays $k$ actions, one in each of the bandit problems, and observes the sum of the losses that correspond with these $k$ actions. 
Then, the set $S$ of actions is given as follows:
\begin{equation}
\label{eq:multitaskset}
S = \left\lbrace 
x \in \{0,1\}^d ~:~ 
\forall j \in [k]~\sum_{i=(j-1)n+1}^{jn} x(i) = 1 \right\rbrace~.
\end{equation}

\begin{theorem}[multitask MAB]
\label{thm:main}
Assume that $n \ge 2$, and let the set of actions $S \subseteq \{0,1\}^d$ be as defined in \cref{eq:multitaskset}. Any learning algorithm for the multitask bandit problem must incur at least~$\tOmega(k^{3/2} \sqrt{dT})$~expected regret in the worst~case.
\end{theorem}

The bound in the theorem hides a factor of $\log^{-1/2} T$ which is an artifact of our construction and is likely to be redundant.
Note, however, that up to logarithmic factors the bound is tight
and matches the upper bounds of \cite{bubeck2012towards} and \cite{hazan2016volumetric}.
%

The lower bound of \cref{thm:main} does not hold for any set $S$ but rather to an instance of the multitask bandit problem. 
However, as we show in the following results, it still is general enough to imply lower bounds for two important instances of bandit linear optimization.
Our next theorem gives a lower bound for the bandit shortest path problem, and shows that even when we limit the action set $S$ to paths in a certain graph, the regret of the learner can still be forced to be large.
Formally, given a connected DAG $G = (V,E)$ with $d$ edges and two nodes $s,t \in V$, we define the set of actions $S \subseteq \{0,1\}^d$ as follows:
\begin{equation}
\label{eq:actionsetshortestpath}
S = \left\lbrace x \in \{0,1\}^d~:~\textrm{the set}~\{e \in E : x(e) = 1\} \textrm{~forms an \stpath path} \right\rbrace~.
\end{equation}
Then, we have the following:

\begin{theorem}[online shortest paths]
\label{corrollary:shortestpath}
Assume that $k \le d/2$. There exists a graph with $d$ edges such that any \stpath path has exactly $k$ edges (see \cref{fig:graph}), for which the action set $S$ is defined as in \cref{eq:actionsetshortestpath}. Against this graph any online learning algorithm for the bandit shortest path problem must suffer at least $\tOmega(k^{3/2} \sqrt{dT})$ expected regret in the worst~case.
\end{theorem}
Again, the theorem implies that the tight regret rate for bandit shortest path is $\wt{\Theta}(k^{3/2}\sqrt{dT})$, contrary to what was conjectured in the literature \citep{cesa2012combinatorial}.

Our last main result shows a lower bound for the online ranking problem. This problem can be cast as finding a maximum matching in the complete bipartite graph $K_{k,n}$, that has $d = kn$ edges. The set of all of these matchings is represented by the action set $S \subseteq \{0,1\}^d$, and the non-zero entries of every $x \in S$ indicate which edges participate in the matching that corresponds with $x$. Formally,
\begin{equation}
\label{eq:actionsetperm}
S = \left\lbrace x \in \{0,1\}^d~:~ \forall j \in [k]~\sum_{i=(j-1)n+1}^{jn} x(i) = 1,~\forall l \in [n]~\sum_{i=1}^k x((i-1)n + l) = 1 \right\rbrace~.
\end{equation}

\begin{theorem}[online ranking]
\label{corrollary:permutations}
Assume that $k \le n/2$. Consider the problem of online ranking between $k$ and $n$ elements, whose action set $S$ is defined in \cref{eq:actionsetperm}. Any bandit learning algorithm for this problem must suffer at least~$\tOmega(k^{3/2} \sqrt{dT})$ expected regret in the worst~case.
\end{theorem}
%

\section{Proofs}

\subsection{Main result}
\label{sec:proofmain}
In this section we prove \cref{thm:main}. We show a lower bound of $\tOmega(k^{3/2} \sqrt{dT})$ on the regret of any online learning algorithm applied to an instance of the multitask bandit problem. 
Surprisingly, the factor $\sqrt{k}$ improvement is obtained via a simple modification of previous constructions~\citep{audibert2014regret}.

We start by applying Yao's minimax principle, implying that it suffices to show randomized strategy for the environment that forces any deterministic learning algorithm to suffer~$\tOmega(k^{3/2} \sqrt{dT})$~regret in expectation. 
We shall construct the environment's strategy as follows. 

Set $\epsilon = \sigma \sqrt{kd / (4T)}$. Before the game begins, the environment chooses the best arm in each of the $k$ problems in $S$ uniformly at random, and denote the vector indicating this choice by~$x^\star \in S$.  At every round~$t$, the environment samples $Z_t \sim \N(0,\sigma^2)$. Denote the loss generated by environment on round $t$ as $L'_t(i) = 1/2 - \epsilon \cdot x^\star(i) + Z_t$ for $i = 1,2,\ldots,d$. 

The idea behind this construction is as follows. 
In order to avoid large losses and minimize her regret, the learner has to identify the best arm in each of the $k$ subproblems, namely, to recover $x^\star$.
Now, suppose that the losses of each coordinate were sampled independently, and each entry in $L_t'$ were to receive an i.i.d.~sample of the Gaussian noise. 
Then the variance of the loss observed by the learner, namely of the random variable $L_t \cdot x$ for any choice of $x \in S$, is of the order of $k$. 
On the other hand, because of the correlation between the losses of the different coordinates in the construction above, the variance of the observed loss is of the order of $k^2$. 
This allows us to gain and additional $\sqrt{k}$ factor in the lower bound on the regret.
Note that crucially, the learner always observes a sum of $k$ random noise terms and can never peek into the individual terms in the sum (this is due to the bandit feedback and the specific structure of the decision set $S$); hence, the correlation in the noise cannot be exploited by the learner and the increase in the overall variance comes at no~price.

For the construction above, we have the following lemma.

\begin{lemma}
\label{lemma:lowerboundgaussian}
Any deterministic player must suffer regret of at least~$\sigma k^{3/2} \sqrt{dT} / 8$ in expectation against an environment that plays the losses $L'_1,\ldots,L'_T$.
\end{lemma} 

To show that \cref{thm:main} holds we need to show that the learner suffers large regret against an environment that plays losses that are bounded in $[0,1]^d$. While the losses we have constructed $L'_1,\ldots,L'_T$ are unbounded, for the right choice of $\sigma$ they are bounded with high probability. We now show that this allows us to obtain a lower bound on the regret against an environment that plays losses $L_1,L_2,\ldots,L_T$, such that $L_t(i) = \clip(L'_t(i))$ for $\clip(a) = \max \{ \min \{a, 1\},0\}$.

\begin{theorem}
\label{thm:lowerboundclip}
Assume that $T \ge k d$ and let $\sigma^2 = 1/(192+96\log T)$. Any deterministic player must suffer an expected regret of at least $\sigma k^{3/2} \sqrt{dT} / 16$ against an environment that plays the losses~$L_1,\ldots,L_T$.
\end{theorem}
The proof of \cref{thm:main} is now given by setting the value of $\sigma$ into the bound in \cref{thm:lowerboundclip}.

\subsection{Bandit shortest path}
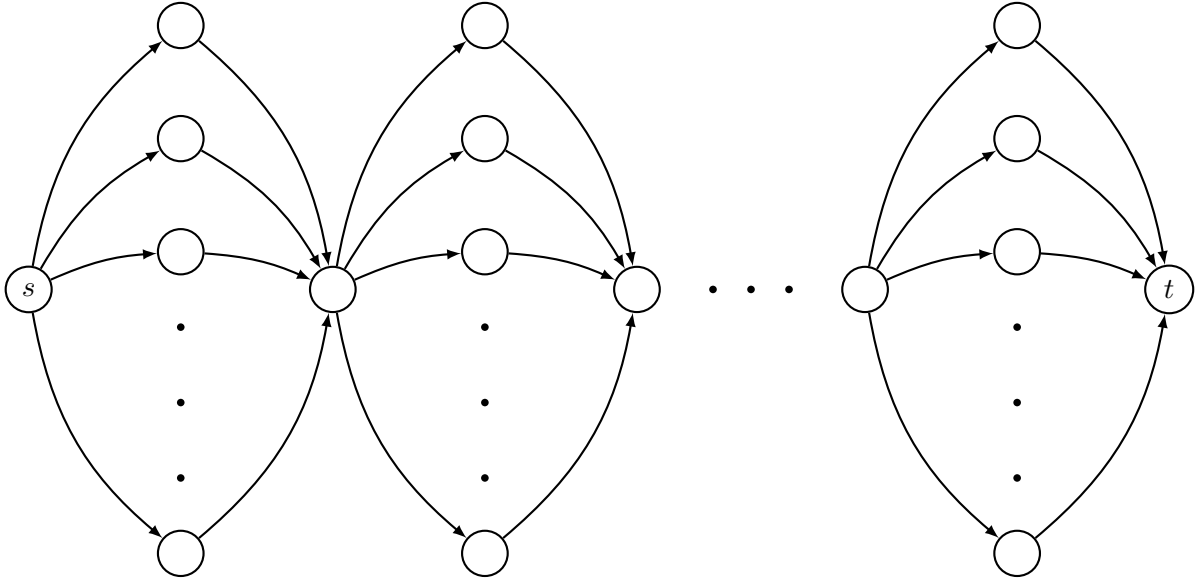
\begin{figure}
\centering
\begin{tikzpicture}[auto,swap]
\begin{scope}[every node/.style={circle, thick, draw, minimum size=0.6cm}]
    \node (A) at (-7.5,0) {$s$};
    \node (B) at (-5.5,3.5) {};
    \node (C) at (-5.5,2) {};
    \node (D) at (-5.5,0.5) {};
    \node (E) at (-5.5,-3.5) {};
    \node (F) at (-3.5,0) {} ;
    \node (G) at (-1.5,3.5) {};
    \node (H) at (-1.5,2) {};
    \node (I) at  (-1.5,0.5) {};
    \node (J) at (-1.5,-3.5) {};
    \node (K) at (0.5,0) {};
    \node (L) at (3.5,0) {};
    \node (M) at (5.5,3.5) {};
    \node (N) at (5.5,2) {};
    \node (O) at (5.5,0.5) {};
    \node (P) at (5.5,-3.5) {};
    \node (Q) at (7.5,0) {$t$};
\end{scope}

\fill (-5.5,-2.5) circle (0.05);
\fill (-5.5,-1.5) circle (0.05);
\fill (-5.5,-0.5) circle (0.05);

\fill (-1.5,-2.5) circle (0.05);
\fill (-1.5,-1.5) circle (0.05);
\fill (-1.5,-0.5) circle (0.05);

\fill (5.5,-2.5) circle (0.05);
\fill (5.5,-1.5) circle (0.05);
\fill (5.5,-0.5) circle (0.05);

\fill (1.5,0) 	circle (0.05);
\fill (2,0) 	circle (0.05);
\fill (2.5,0) 	circle (0.05);

\begin{scope}[>={latex[black]},
              every node/.style={}, 
              every edge/.style={draw=black,thick}]
    \path [->] (A) edge[bend left=20]  (B);
    \path [->] (A) edge[bend left=15]  (C);
    \path [->] (A) edge[bend left=10]  (D);
    \path [->] (A) edge[bend right=20] (E);
    
    \path [->] (B) edge[bend left=20]  (F);
    \path [->] (C) edge[bend left=15]  (F);
    \path [->] (D) edge[bend left=10]  (F); 
    \path [->] (E) edge[bend right=20] (F); 
    
    \path [->] (F) edge[bend left=20]  (G);
    \path [->] (F) edge[bend left=15]  (H);
    \path [->] (F) edge[bend left=10]  (I);
    \path [->] (F) edge[bend right=20] (J);
    
    \path [->] (G) edge[bend left=20]  (K);
    \path [->] (H) edge[bend left=15]  (K);
    \path [->] (I) edge[bend left=10]  (K); 
    \path [->] (J) edge[bend right=20] (K); 
    
    \path [->] (L) edge[bend left=20]  (M);
    \path [->] (L) edge[bend left=15]  (N);
    \path [->] (L) edge[bend left=10]  (O);
    \path [->] (L) edge[bend right=20] (P);
    
    \path [->] (M) edge[bend left=20] (Q);
    \path [->] (N) edge[bend left=15] (Q);
    \path [->] (O) edge[bend left=10] (Q); 
    \path [->] (P) edge[bend right=20] (Q); 
\end{scope}
\end{tikzpicture}
\caption{Graph for the lower bound. The graph consists of $k/2$ layers, in each the learner has to choose one of $d/k$ vertices for an \stpath path to pass through.\label{fig:graph}}
\end{figure}
In this section we show a lower bound for the bandit shortest path problem, proving \cref{corrollary:shortestpath}. 
Suppose without loss of generality that $k$ and $d$ are even, and that $d$ is a multiple of $k$. We show a lower bound on the regret by constructing a graph that simulates the multitask bandit problem with $k/2$ problems of $d/k$ arms each.

This graph is shown in \cref{fig:graph}. The graph consists of $d$ edges and $d/2+k/2+1$ vertices set in $k/2$ layers. Each layer has an incoming vertex connected to $d/k$ intermediate vertices, all of them connected to the same outgoing vertex. This outgoing vertex is the incoming vertex of the next layer and so forth. Note that to form an \stpath path the learner has to pass through exactly one of the $d/k$ vertices in each layer, and therefore every such path has exactly $k$ edges.

Now, given the losses $L_1,L_2,\ldots,L_T$ generated by the environment of \cref{sec:proofmain}, we shall construct an environment for the shortest path problem such that the regret of the learner would be the same as the one in the proof of \cref{thm:main}. Indeed, recall that the loss at coordinates $(j-1)d/k+1,\ldots,jd/k$ is associated with the losses of the $j$'th $d/k$-armed bandit problem. 
Then on round $t$ for the $j$'th layer of the graph, we can set the losses $L_t((j-1)d/k+1),\ldots,L_t(jd/k)$ to the edges going from the incoming vertex to the intermediate vertices, and a loss of $0$ to the edges going from the intermediate vertices to the outgoing vertex.

Therefore, we have a bijection between any \stpath path and a set of $k/2$ arms in the aforementioned multitask bandit problem, such that the sum of the losses on the edges of the \stpath path and the sum of the losses of these $k/2$ arms are the same. We conclude by invoking \cref{thm:main} that says that any learner must suffer an expected regret of at least $\tOmega(k^{3/2} \sqrt{d T})$, as claimed.

\subsection{Online ranking}

In this section we prove \cref{corrollary:permutations} by a similar construction to the one in \cref{sec:proofmain}, for which we present the following random environment.

Set $\epsilon = \sigma \sqrt{kd / (8T)}$. Before the game starts, the environment samples a maximum matching in $K_{k,n}$ unfiromly at random, and denote the vector indicating this choice by $x^\star \in S$, for the set $S$ defined in \cref{eq:actionsetperm}. At every round~$t$, the environment samples $Z_t \sim \N(0,\sigma^2)$. Denote the loss  generated by the environment on round $t$ as $L'_t(i) = 1/2 - \epsilon \cdot x^\star(i) + Z_t$ for all $i = 1,2,\ldots,d$.

We have the following lemma.

\begin{lemma}
\label{lemma:gaussianperm}
Any deterministic player must suffer regret of at least~$\sigma k^{3/2} \sqrt{dT} / 8$ in expectation against an environment that plays the losses $L'_1,\ldots,L'_T$.
\end{lemma} 

Now to prove \cref{corrollary:permutations}, the result above can be adapted to bounded losses in the same manner as done in \cref{thm:lowerboundclip}.

\section{Additional proofs}

\subsection{Proof of \cref{lemma:lowerboundgaussian}}

\begin{proof}
Let us denote by $i^\star_1,\ldots,i^\star_k$ the locations of the non-zero coordinates of the random variable $x^\star$, arranged in increasing order.
We next introduce the random variables $T_1,\ldots,T_k$, where each $T_j$ is the number of times the learner played an $x_t$ such that $x_t(i^\star_j) = 1$. 
For each $x \in S$, we introduce the notations $\Pr_x$ and $\E_x$ indicating probability and expectation with respect to the marginal distributions under which $x^\star = x$. Then,

\begin{align}
\Regret &= \E \left[ \sum_{t=1}^T L'_t \cdot x_t - \min_{x \in S} \sum_{t=1}^T L'_t \cdot x \right] \nonumber \\
&\ge \E \left[ \sum_{t=1}^T L'_t \cdot x_t - \sum_{t=1}^T L'_t \cdot x^\star \right] \nonumber \\
&= \frac{1}{n^k} \sum_{x \in S} \E_x \left[ \sum_{t=1}^T L'_t \cdot x_t - \sum_{t=1}^T L'_t \cdot x \right] \nonumber \\
&= \frac{1}{n^k} \sum_{x \in S} \epsilon \cdot \E_x \left[\sum_{j=1}^k \left( T - T_j \right) \right] \nonumber \\ 
&= \epsilon \left(kT - \sum_{j=1}^k \frac{1}{n^k} \sum_{x \in S} \E_x \left[T_j \right] \right)~,
\label{eq:lbregret}
\end{align}
and in order to proceed, we need to upper bound $\E_x [T_j]$ for each $j$.

For every $x \in S$ and $j \in [k]$ we introduce a new distribution, which is the same as $\Pr_x$ except that the loss of coordinate $i^\star_j$ is also $1/2+Z_t$. We shall refer to these new laws by $\Pr_{x,-j}$ and $\E_{x,-j}$. 
Let~$\lambda_t$~be the loss observed at time $t$, and $\lambda^{(t)} = (\lambda_1,\ldots,\lambda_t)$ be the losses observed up to and including time $t$. Then, since the sequence $\lambda^{(T)}$ determines the actions of the learner over the entire game, and by Pinsker's inequality,
\begin{align}
\E_x[T_j] - \E_{x,-j}[T_j] &\le T \cdot \TV{\Pr_{x,-j} \left[\lambda^{(T)} \right]}{\Pr_{x} \left[\lambda^{(T)}\right]} \nonumber \\
&\le T \sqrt{\half \KL{\Pr_{x,-j} \left[\lambda^{(T)} \right]}{\Pr_{x} \left[\lambda^{(T)}\right]} }~.
\label{eq:tjcomparison}
\end{align}
Moreover, by the chain rule of KL-divergence, $\KL{\Pr_{x,-j} [\lambda^{(T)}]}{\Pr_{x} [\lambda^{(T)}]}$ equals 
\begin{equation}
\sum_{t=1}^T \E_{\lambda^{(t-1)} \sim \Pr_{x,-j}} \left[ \KL{\Pr_{x,-j} \left[\lambda_t \Bigr| \lambda^{(t-1)} \right]}{\Pr_{x} \left[\lambda_t \Bigr| \lambda^{(t-1)}\right]} \right]~.
\label{eq:klchainrule}
\end{equation}
Consider a single term in the sum, and recall that $\lambda^{(t-1)}$ determines the action $x_t$ chosen by the learner on round $t$. If $x_t(i^\star_j) = 0$, the loss observed under $\Pr_x$ and $\Pr_{x,-j}$ are the same, and the KL divergence is 0. If $x_t(i^\star_j) = 1$ then the observed losses under $\Pr_x$ and $\Pr_{x,-j}$ are both Gaussian whose means are $\epsilon$ apart, and the variance of both of them is $\sigma^2 k^2$. 
Therefore, 
\[
\KL{\Pr_{x,-j} \left[\lambda_t \Bigr| \lambda^{(t-1)} \right]}{\Pr_{x} \left[\lambda_t \Bigr| \lambda^{(t-1)}\right]} \le \frac{\epsilon^2}{2 k^2 \sigma^2}~.
\]
Plugging the above back into \cref{eq:klchainrule},
\[
\KL{\Pr_{x,-j} \left[\lambda^{(T)} \right]}{\Pr_{x} \left[\lambda^{(T)}\right]} \le \sum_{t=1}^T \Pr_{x,-j} \left[ x_t(i^\star_j) = 1 \right] \cdot \frac{\epsilon^2}{2k^2 \sigma^2} = \frac{\epsilon^2}{2k^2 \sigma^2} \E_{x,-j} [T_j]~,
\]
and the latter back into \cref{eq:tjcomparison}, we get $\E_x[T_j] \le \E_{x,-j}[T_j] + \epsilon T / (2 k \sigma) \cdot \sqrt{\E_{x,-j} \left[T_j \right]}$.

Next, we need the following lemma that we prove on \cref{sec:proofaveragetjlemma}.

\begin{lemma}
\label{lemma:onaveragetj}
In the conditions of \cref{lemma:lowerboundgaussian} and by the construction above, we have
\[
\frac{1}{n^k} \sum_{x \in S} \E_{x,-j} \left[T_j \right] = \frac{T}{n}~.
\]
\end{lemma}

Note that $n \ge 2$ by assumption. Therefore, for all $j = 1,2,\ldots,k$,
\begin{align*}
\frac{1}{n^k} \sum_{x \in S} \E_x [T_j] &\le \frac{1}{n^k} \sum_{x \in S} \E_{x,-j}[T_j] + \frac{\epsilon T}{2 k \sigma} \cdot \frac{1}{n^k} \sum_{x \in S} \sqrt{\E_{x,-j} [T_j]} \\
&\le \frac{1}{n^k} \sum_{x \in S} \E_{x,-j}[T_j] + \frac{\epsilon T}{2 k \sigma} \sqrt{ \frac{1}{n^k} \sum_{x \in S} \E_{x,-j} [T_j]} \\
&\le \frac{T}{2} + \frac{\epsilon T}{2 \sigma} \sqrt{ \frac{T}{kd}}~,
\end{align*}
since $d = k  n$.
Let us now return to \cref{eq:lbregret}. We can lower bound the regret as
\begin{align*}
\Regret &\ge \epsilon \left( kT - \sum_{j=1}^k \left( \frac{T}{2} + \frac{\epsilon T}{2 \sigma} \sqrt{\frac{T}{k d}} \right) \right) \\
&= \epsilon k T \left( \half - \frac{\epsilon}{2 \sigma} \sqrt{\frac{T}{k d}} \right)~.
\end{align*}
For our choice of $\epsilon$, we get that $\epsilon / (2 \sigma) \sqrt{T / (k d)}$ is at most $1/4$, and so
\[
\Regret \ge \sigma \sqrt{\frac{k d}{4T}} \cdot k T \left( \half - \frac{1}{4} \right) = \frac{\sigma}{8} k^{3/2} \sqrt{d T}~,
\]
as claimed. 
\end{proof}

\subsection{Proof of \cref{lemma:onaveragetj}}
\label{sec:proofaveragetjlemma}

\begin{proof}
For any choice $i^\star_1,i^\star_2,\ldots,i^\star_k$, let us denote by $x(i^\star)$ the corresponding $x^\star \in S$. 
Following \cite{audibert2014regret}, we consider
\[
\sum_{x \in S} \E_{x,-j} \left[T_j \right] = \sum_{i^\star_1,\ldots,i^\star_{j-1},i^\star_{j+1},\ldots,i^\star_k} \sum_{i^\star_j} \E_{x(i^\star),-j} \left[T_j \right]~.
\]

Now, keeping $i^\star_1,\ldots,i^\star_{j-1},i^\star_{j+1},\ldots,i^\star_k$ fixed the distribution $\Pr_{x(i^\star),-j}$ is the same for any choice of $i^\star_j$ and therefore, since at every round of the game the learner must choose exactly one arm in the $j$'th problem, we must have $\sum_{i^\star_j} \E_{x(i^\star),-j} [T_j] = T$.

Putting it all together, we obtain
\[
\sum_{i^\star_1,\ldots,i^\star_{j-1},i^\star_{j+1},\ldots,i^\star_k} \sum_{i^\star_j} \E_{x(i^\star),-j} \left[T_j \right] = \sum_{i^\star_1,\ldots,i^\star_{j-1},i^\star_{j+1},\ldots,i^\star_k} T = n^{k-1} T~,
\]
and thus
\[
\frac{1}{n^k} \sum_{x \in S} \E_{x,-j} \left[T_j \right] = \frac{1}{n^k} n^{k-1} T = \frac{T}{n}~. \qedhere
\]
\end{proof}

\subsection{Proof of \cref{thm:lowerboundclip}}

\begin{proof}
Let $X_1,X_2,\ldots,X_T$ be the predictions of the learner against an environment that plays $L_1,L_2,\ldots,L_T$, and let $\hat{\mathrm{R}}_T$ be the regret attained by the learner,
\[
\hat{\mathrm{R}}_T = \sum_{t=1}^T L_t \cdot X_t - \min_{x \in S} \sum_{t=1}^T L_t \cdot x~.
\]
Also define the pretend-regret obtained by playing $X_1,X_2,\ldots,X_T$ against an enivronment that plays~$L'_1,L'_2,\ldots,L'_T$ as
\[
\hat{\mathrm{R}}'_T = \sum_{t=1}^T L'_t \cdot X_t - \min_{x \in S} \sum_{t=1}^T L'_t \cdot x~.
\]

Now note that if it happens that at every round $t$, all coordinates of $L'_t$ are between 0 and 1, then $\hat{\mathrm{R}}_T = \hat{\mathrm{R}}'_T$. Denote this event by $E$. Then,
\begin{equation}
\label{eq:regretclipbound}
\E[\hat{\mathrm{R}}'_T] \le \E[\hat{\mathrm{R}}_T] + kT \cdot \Pr[E^c] = \Regret + kT \cdot \Pr[E^c]
\end{equation}
where the inequality is true since the regret is at most $kT$ with probability 1.

It thus remains to upper bound the probability that $E$ does not occur. We will show that~$\Pr[E^c] \le \epsilon / 8$, which by combining \cref{eq:regretclipbound} and \cref{lemma:lowerboundgaussian} would yield:
\[
\Regret \ge \frac{\sigma k^{3/2} \sqrt{dT}}{8} - \frac{\sigma k^{3/2} \sqrt{dT}}{16} = \frac{\sigma k^{3/2} \sqrt{dT}}{16}~,
\]
as required.
Now, for $E$ to occur it suffices that $\epsilon \le 1/4$ and that $Z_t \le 1/4$ for every round $t$. 
Since
\[
\epsilon = \sqrt{\frac{\sigma^2 kd}{4T}} \le \sqrt{\frac{kd}{(192+96 \log T) T}} \le \sqrt{\frac{1}{192}} \le \frac{1}{4}~,
\]
by our choice of $\epsilon$ and $\sigma$ and since $T \ge kd$ by assumption, 
we have that the probability $\Pr[E^c]$ is upper bounded by the probability that $Z_t > 1/4$ at some (at least one) round $t$. 
Employing the standard tail bound $\Pr(Z > x) \le \exp(-x^2/2\sigma^2)$ for the normal distribution and the union bound, the latter is bounded by
\begin{align*}
T \cdot \Pr[Z_1 > 1/4] &\le T \exp \left( - \frac{1}{2 \sigma^2} \left( \frac{1}{4} \right)^2 \right) \\
&= T \exp \left( -(6 + 3 \log T) \right) \\
&= e^{-6} \frac{1}{T^2}~.
\end{align*}
Therefore, for the probability that $E$ does not occur to be at most $\epsilon / 8$ it suffices to have 
\[
16 e^{-6} \frac{1}{T^2} \le \sqrt{\frac{1}{(192+96\log T) T}}~.
\]
Rearranging the terms it then suffices to have $T^3 \ge 0.16 + 0.08 \log T$, that holds for any~$T \ge 1$. \qedhere
\end{proof}

\subsection{Proof of \cref{lemma:gaussianperm}}

\begin{proof}
Let us denote by $i^\star_1,\ldots,i^\star_k$ the locations of the nonzero coordinates of the random variable $x^\star$, arranged in increasing order.
We next introduce the random variables $T_1,\ldots,T_k$, where each $T_j$ is the number of times the learner played an $x_t$ such that $x_t(i^\star_j) = 1$. 
For each $x \in S$, we introduce the notations $\Pr_x$ and $\E_x$ indicating probability and expectation with respect to the marginal distributions under which $x^\star = x$. Then,
\begin{align}
\Regret &= \E \left[ \sum_{t=1}^T L'_t \cdot x_t - \min_{x \in S} \sum_{t=1}^T L'_t \cdot x \right] \nonumber \\
&\ge \epsilon \left(kT - \sum_{j=1}^k \frac{(n-k)!}{n!} \sum_{x \in S} \E_x \left[T_j \right] \right)~, \label{eq:regretperm}
\end{align}
and in order to proceed, we need to upper bound $\E_x [T_j]$ for each $j$.

For every $x \in S$ and $j \in [k]$ we introduce a new distribution, which is the same as $\Pr_x$ except that the loss of coordinate $i^\star_j$ is also $1/2+Z_t$. We shall refer to these new laws by $\Pr_{x,-j}$ and~$\E_{x,-j}$. 
From now on the proof proceeds similarly to that of \cref{lemma:lowerboundgaussian}, with the exception that \cref{lemma:onaveragetj} is replaced by the following lemma, whose proof can be found in \cref{sec:proofpermaveragetj}.
\begin{lemma}
\label{lemma:onaveragetjperm}
In the conditions of \cref{lemma:gaussianperm} and by the construction above, we have
\[
\frac{(n-k)!}{n!} \sum_{x \in S} \E_{x,-j} \left[T_j \right] \le \frac{T}{n-k+1}~.
\]
\end{lemma}

Recall that $k \le n/2$ by assumption, that in particular implies $n-k+1 \ge 2$ as well as $n-k+1 \ge n/2$. Therefore, for all $j = 1,2,\ldots,k$,
\begin{align*}
\frac{(n-k)!}{n!} \sum_{x \in S} \E_x [T_j] &\le \frac{(n-k)!}{n!} \sum_{x \in S} \E_{x,-j}[T_j] + \frac{\epsilon T}{2 k \sigma} \cdot \frac{(n-k)!}{n!} \sum_{x \in S} \sqrt{\E_{x,-j} [T_j]} \\
&\le \frac{(n-k)!}{n!} \sum_{x \in S} \E_{x,-j}[T_j] + \frac{\epsilon T}{2 k \sigma} \sqrt{ \frac{(n-k)!}{n!} \sum_{x \in S} \E_{x,-j} [T_j]} \\
&\le \frac{T}{2} + \frac{\epsilon T}{2 k \sigma} \sqrt{ \frac{2T}{n}}~.
\end{align*}
Let us now return to \cref{eq:regretperm}. Using the fact that $n = d/k$, we can lower bound the regret as
\[
\Regret \ge \epsilon k T \left( \half - \frac{\epsilon}{\sigma} \sqrt{\frac{T}{2k d}} \right)~,
\]
which, by our choice of $\epsilon$, allows us to obtain the desired lower bound.
\end{proof}

\subsection{Proof of \cref{lemma:onaveragetjperm}}
\label{sec:proofpermaveragetj}

\begin{proof}
Recall that we sample $x^\star$ uniformly at random from $S$, the set defined in \cref{eq:actionsetperm}, and denote by $U(S)$ the uniform distribution over $S$. Then, recalling the random variables $i^\star_1,i^\star_2,\ldots,i^\star_k$ we can compute
\[
\E_{x^\star \sim U(S)} \E_{x^\star, -j} \left[ T_j \right]~,
\]
by conditioning on $i^\star_1,\ldots,i^\star_{j-1},i^\star_{j+1},\ldots,i^\star_k$ and taking the outer expectation only over $i^\star_j$.

Now, there are exactly $n-k+1$ possible ways to choose $i^\star_j$ in order to complete a maximal matching. In addition, the distribution $\Pr_{x^\star,-j}$ is the same for any possible choice of $i^\star_j$, and since at every round of the game the learner must choose exactly one position for the $j$'th element, we must have 
\begin{align*}
&\E_{x^\star \sim U(S)} \left[ \E_{x^\star,-j} \left[T_j \right]~\Big|~i^\star_1 = i_1,\ldots,i^\star_{j-1} = i_{j-1},i^\star_{j+1} = i_{j+1},\ldots,i^\star_k = i_k \right] \\
&\qquad = \frac{1}{n-k+1} \sum_{x^\star \in S} \textbf{1}_{\left[i^\star_1 = i_1,\ldots,i^\star_{j-1} = i_{j-1},i^\star_{j+1} = i_{j+1},\ldots,i^\star_k = i_k \right]} \E_{x^\star,-j} [T_j] \\
&\qquad \le \frac{T}{n-k+1}~. \qedhere
\end{align*}
\end{proof}

\section{Conclusion and open problems}

In this paper, we gave a tight characterization of the optimal regret rate in bandit combinatorial optimization and proved that it grows as $\wt{\Theta}(k^{3/2}\sqrt{dT})$, disproving the conjectures of \cite{cesa2012combinatorial} and \cite{audibert2014regret}.
Our lower bounds apply to important instances of the framework, including the bandit versions of the online shortest path and the online ranking problems.

An interesting direction for future work is to explore instance-specific bounds, i.e., bounds that depend on the structure of the specific action set $S$ used by the learner. 
What are the geometric and combinatorial properties of the set $S$ that dictate the optimal rate of regret in the induces learning problem?
In particular, in the specific context of the bandit shortest path problem, what are the graph-theoretic properties of the network that govern the difficulty of the online problem?
Even in extremely simple graphs, such as the two-dimensional directed grid over $n^2$ nodes (where the $s$ and $t$ nodes are located in two opposite corners), characterizing the optimal rate of regret remains an open problem.
We suspect such problems to be non-trivial already in full-information online combinatorial optimization, but expect the bandit setting to be particularly  challenging.

For the problem of online ranking, \cref{corrollary:permutations} handles the case of $k \times n$ permutations in which $k$ is smaller than $n$. However, quantifying the rate of regret in the important case of full  permutations (i.e., with $k=n$) remains an open problem. In particular, is the optimal regret~$\Theta(n^2 \sqrt{T})$ in this~setting? 


\bibliographystyle{abbrvnat}
\bibliography{bib}

\end{document}